\def\year{2022}\relax
\documentclass[letterpaper]{article} 
\usepackage{aaai22}  
\usepackage{times}  
\usepackage{helvet}  
\usepackage{courier}  
\usepackage[hyphens]{url}  
\usepackage{graphicx} 
\usepackage{natbib}  
\usepackage{caption} 
\DeclareCaptionStyle{ruled}{labelfont=normalfont,labelsep=colon,strut=off} 
\usepackage{algorithm}
\usepackage{algorithmic}
\usepackage{color,xcolor}
\usepackage{multirow}
\usepackage{graphicx}
\usepackage{amsmath}
\usepackage{booktabs}
\usepackage{enumitem}

\usepackage{newfloat}
\usepackage{listings}
\floatstyle{ruled}
\newfloat{listing}{tb}{lst}{}
\floatname{listing}{Listing}
\title{Small Changes Make Big Differences: Improving Multi-turn Response Selection \\in Dialogue Systems via Fine-Grained Contrastive Learning}
\author{
Yuntao Li\textsuperscript{1},
Can Xu\textsuperscript{3},
Huang Hu\textsuperscript{3},
Lei Sha\textsuperscript{2},
Yan Zhang\textsuperscript{1},
Daxin Jiang\textsuperscript{3}\thanks{Corresponding author: Daxin Jiang (djiang@microsoft.com).}\\
}
\affiliations{
\textsuperscript{1}{Department of Machine Intelligence, Peking University, Beijing, China}\\
\textsuperscript{2}{Department of Computer Science, University of Oxford, United Kingdom}\\
\textsuperscript{3}{Microsoft STCA, Beijing, China}\\
\textsuperscript{1}\{li.yt, zhyzhy001\}@pku.edu.cn; 
\textsuperscript{2}\{lei.sha\}@cs.ox.ac.uk;
\textsuperscript{3}\{caxu, huahu, djiang\}@microsoft.com
}

\usepackage{bibentry}

\begin{document}

\maketitle

\begin{abstract}
Retrieve-based dialogue response selection aims to find a proper response from a candidate set given a multi-turn context. Pre-trained language models (PLMs) based methods have yielded significant improvements on this task. The sequence representation plays a key role in the learning of matching degree between the dialogue context and the response. However, we observe that different context-response pairs sharing the same context always have a greater similarity in the sequence representations calculated by PLMs, which makes it hard to distinguish positive responses from negative ones. Motivated by this, we propose a novel \textbf{F}ine-\textbf{G}rained \textbf{C}ontrastive (FGC) learning method for the response selection task based on PLMs. This FGC learning strategy helps PLMs to generate more distinguishable matching representations of each dialogue at fine grains, and further make better predictions on choosing positive responses. Empirical studies on two benchmark datasets demonstrate that the proposed FGC learning method can generally and significantly improve the model performance of existing PLM-based matching models.\footnote{We will make the code public available later to facilitate reproducing the results.}
\end{abstract}

\section{Introduction}

Building an intelligent conversational agent that can naturally and consistently converse with humans has drawn considerable attention in the field of natural language processing (NLP). So far, there are two kinds of approaches for implementing a dialogue system: generation-based \cite{serban2016building,vinyals2015neural} and retrieval-based methods \cite{lowe2015ubuntu,wu2016sequential,zhou2018multi}. 
In this work, we focus on the problem of multi-turn response selection for retrieval-based dialogue systems. 

Multi-turn response selection is the task of predicting the most proper response using a retrieval model by measuring the matching degree between a multi-turn dialogue context and a set of response candidates. 
Most recently, pre-trained language models (PLMs) have achieved substantial performance improvements in multi-turn response selection \cite{lu2020improving,gu2020speaker,humeau2019poly}. PLM-based models take a pair of context and response as a whole and represent them as a single sequence representation (i.e., context embedding at the position of the [CLS] token). This sequence representation is also regarded as a matching representation, which is further used to decide a score indicating the matching degree between the dialogue context and the response. By further post-training PLMs with in-domain data and auxiliary self-supervised tasks \cite{whang2020domain,whang2020response,xu2020learning}, PLMs are incorporated with in-domain knowledge and achieve state-of-the-art results on benchmarks.

\begin{figure}[htbp]
    \centering
    \includegraphics[width=0.48\textwidth]{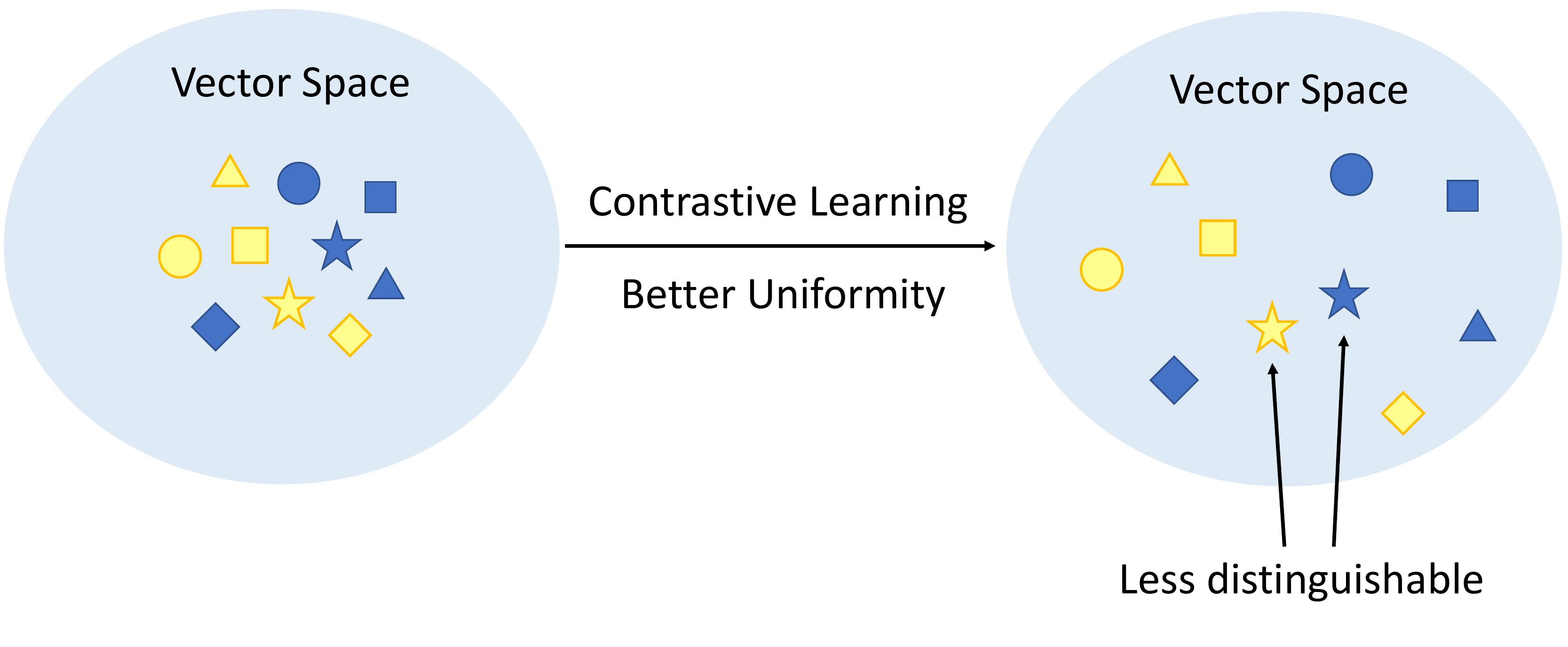}
    \caption{Although a standard contrastive learning can provide better uniformity on the vector space, it is still hard to distinguish between positive (yellow) and negative (blue) dialogues with the same context (in the same shape, e.g., star).}
    \label{fig:contras}
\end{figure}


Despite the success of PLM-based matching models and their various variants, recent studies reveal that the contextualized word and sentence representations in PLM-based models are anisotropic, i.e., representations occupy a narrow cone in the vector space, which largely limits the expressiveness of representations \cite{ethayarajh-2019-contextual,li-etal-2020-sentence}. Contrastive learning provides a way to solve this problem, which uniforms representations over a hypersphere space as pointed out by \citet{wang2020understanding}. Although employing the standard constrative learning method \cite{chen2020simple,fang2020cert} with in-batch negative sampling enhances the uniformity of representations, we further observe by experiments (Section \ref{exp:simple_contrast}) that the matching representation of two dialogues with the same context but different responses are too similar, as is shown in Figure \ref{fig:contras}. This is mainly due to the following two reasons: (1) The overlap of the same context tokens in different context-response pairs makes the matching representation highly similar since the representation is aggregated from all tokens in the context-response pair. (2) The in-batch negative samples are highly likely to be different in both context and response. This phenomenon makes the representations less distinguishable and makes it hard to tell dialogues with positive responses from negative ones.

To address the aforementioned issues, we propose a \textbf{F}ine-\textbf{G}rained \textbf{C}ontrastive learning (FGC) approach to fine-tune matching representations for the response selection task. FGC introduces the connections between each dialogue pair with the same context and different responses into the training process with contrastive learning. In contrast to the standard contrastive learning method, which takes every other dialogue as negative examples, our proposed FGC takes context and response as separate parts and focuses on distinguishing between positive and negative dialogues with the same context. During fine-grained contrastive learning, each dialogue is converted into an augmented dialogue via rule-based transformation on the response utterance. Each dialogue is asked to be close to its augmentation, while the one with a positive response should be far away from the one with a negative response. FGC works totally in a self-supervised way that no additional supervision is required besides the classification label used for response selection training.

We conduct experiments on two response selection benchmarks: the Ubuntu Dialogue Corpus~\cite{lowe2015ubuntu} and the Douban Corpus~\cite{wu2016sequential}. These two datasets have a large variety of both topics and languages. Moreover, our proposed learning framework is independent of the choice of PLMs-based matching models. Therefore, for a comprehensive evaluation, we test FGC with five representative PLMs-based matching models, including the state-of-the-art one brought by self-supervised learning. Our empirical results demonstrate that FGC is able to consistently improve PLMs by up to 3.2\% absolute improvement with an average of 1.7\% absolute improvement in terms of R$_{10}$@1. Besides, We also compare our method with standard-contrastive-learning-enhanced PLMs, which demonstrates the effectiveness of our proposed fine-grained contrastive objective. 

In summary, our contributions in the paper are three-fold:
\begin{itemize}
    \item We propose FGC, a novel fine-grained contrastive learning method, which helps generate better representations of dialogues and improves the response selection task.
    \item FGC shows good generality of effectiveness with various pre-trained language models for enhancing performance.
    \item Experimental results on two benchmark datasets demonstrate that FGC can significantly improve the performance of various strong PLM-based matching models, including state-of-the-art ones.
\end{itemize}

\section{Related Work}
\subsection{Multi-Turn Response Selection}
Earlier studies on retrieval-based response selection focused on single-turn response selection \cite{wang2013dataset,hu2015convolutional,wang2015syntax}. Recently, researchers have paid more attention to the multi-turn response selection. The dual LSTM model \cite{lowe2015ubuntu} was proposed to match the dialog context and response using a dual RNN-based architecture. \citet{zhou2016multi} proposed the multi-view model that encodes dialogue context and response both on the word-level and utterance-level. However, these methods ignore relationships among utterances by simply concatenating the utterances together or converting the whole context to a vector. To alleviate this, \citet{wu2016sequential} proposed the sequential matching network that each utterance in the context first interacts with the response candidate, and then the matching features are aggregated according to the sequential order of multi-turn context. With the rise of the self-attention mechanism \cite{vaswani2017attention}, some studies \cite{zhou2018multi,tao2019multi} explored how to enhance representations with it. Besides, \citet{yuan2019multi} recently revealed that previous methods construct dialogue contextual representation using too much context, which damages the matching effect. They proposed a multi-hop selector to select the relevant utterances in the dialogue context for response matching.

Most recently, pre-trained language models (e.g., BERT \cite{devlin2018bert} and RoBERTa \cite{liu2019roberta}) have shown an impressive performance in the response selection. The post-training method, which helps transfer the representations of BERT from the general domain to the dialogue domain, was proposed by \citet{whang2020domain} and obtained state-of-the-art results. Subsequent researches \cite{gu2020speaker,lu2020improving} focused on incorporating speaker information into BERT and showed its effectiveness in multi-turn response selection. Further, self-supervised learning contributed to the success of pre-trained language models was also applied in several NLP downstream tasks, such as summarization \cite{wang2019self} and the response generation \cite{zhao2020learning}. In the task of response selection, \citet{whang2020response} and \citet{xu2020learning} indicated that incorporating well-designed self-supervised tasks according to the characteristics of the dialogue data into BERT fine-tuning can help with the multi-turn response selection. \citet{han2021fine} proposed a fine-grained post-training method for enhancing the pre-trained language model, while the post-training process is computationally expensive than fine-tuning a classification model. \citet{su2020dialogue} proposed a hierarchical curriculum learning framework for improving response selection with PLMs.

\subsection{Contrastive Learning for NLP}
There have been several investigations for contrastive loss formulations for deep neural models, primarily in the computer vision domain. \citet{oord2018representation} proposed a framework for contrastive learning to learn visual representations based on contrastive predictive coding, which predicts the features in latent space by using powerful autoregressive models. \citet{khosla2020supervised} investigated supervised contrastive learning, allowing to leverage label information effectively. Following this trend, some researchers verified the effectiveness of constructive learning in specific NLP tasks. \citet{fang2020cert} proposed pre-training language representation models with a contrastive self-supervised learning objective at the sentence level, outperforming previous methods on a subset of GLUE tasks. \citet{gunel2020supervised} combined the cross-entropy with a supervised contrastive learning objective, showing improvements over fine-tuning RoBERTa-Large on multiple datasets of the GLUE benchmark. Our work differs from previous works in that we do not directly make contrast on one dialogue with all the other dialogues, as the granularity of negative samples constructed using this approach is too coarse to provide sufficient discrimination with the positive ones.

\section{Background}

\subsection{Task Formalization}

The response selection task is to select the best candidate to respond a given multi-turn dialogue context from a pool of candidate responses. Suppose that we have a dataset $D=\{c_i,r_i,y_i\}_{i=1}^{N}$, where $c_i=\{u_i^1,\cdots,u_i^{n_i}\}$ is a multi-turn dialogue context with $n_i$ turns, $r_i$ denotes a candidate response, and $y_i\in \{0,1\}$ denotes a label with $y_i=1$ indicating $r_i$ a proper response for $c_i$ and otherwise $y_i=0$. Our goal is to estimate a matching model $y=f(\cdot,\cdot)$ from $D$. For any given context-response pair $(c,r)$, $f(c,r)$ returns a score that reflects the matching degree between $c$ and $r$.

\subsection{Pre-trained Language Model for Response Selection}

As a trend in these years, pre-trained language models, e.g., BERT\cite{devlin2018bert}, have been widely studied and adapted into numerous NLP tasks, showing several state-of-the-art results. Dialogue response selection is one of them. 

Applying a pre-trained language model into response selection usually involves two steps. The first step is to make domain-adaptive post-training, which continues to train a standard pre-trained language model with a domain-specific corpus. This step helps to transfer the original pre-trained language model into the target domain. 

The second step is to fine-tune the post-trained model with the response selection task. Given a context $c=\{u_1, \cdots, u_m\}$ where $u_t$ is the t-th turn of the dialog context, and a response $r$, the model is asked to predict a score $\hat{y}$ to represent the matching degree between $c$ and $r$. To achieve this, a special token [EOT] is added at the end of each turn to distinguish them in the context $c$. 
Utterances from both the context $c$ and response $r$ are concatenated with separator [EOT] and [SEP] between them.
Taking $x$ as input, BERT returns a sequence of vectors with the same length as $x$. The output of the first place $\mathbf{s}_{\texttt{[CLS]}}$ is an aggregated representation vector that holds the information of interaction between context $c$ and response $r$. A relevance score $\hat{y}$ is computed based on $\mathbf{s}_{\texttt{[CLS]}}$ and optimized through a binary classification loss. 
\begin{equation}
    \begin{aligned}
    \hat{y} &= \sigma ( \mathbf{W}_{sel} \mathbf{s}_{\texttt{[CLS]}} + b ) \\
    \mathcal{L}_{\texttt{sel}} &= - \left(  y \log \hat{y} + (1 - y) \log (1 - \hat{y}) \right),
    \end{aligned}
\end{equation} where $\mathbf{W}_{sel}$ and $b$ are parameters 


\section{Methodology}

\subsection{Overview}

\begin{figure*}[htbp]
    \centering
    \includegraphics[width=0.85\textwidth,trim=0 0 0 0,clip]{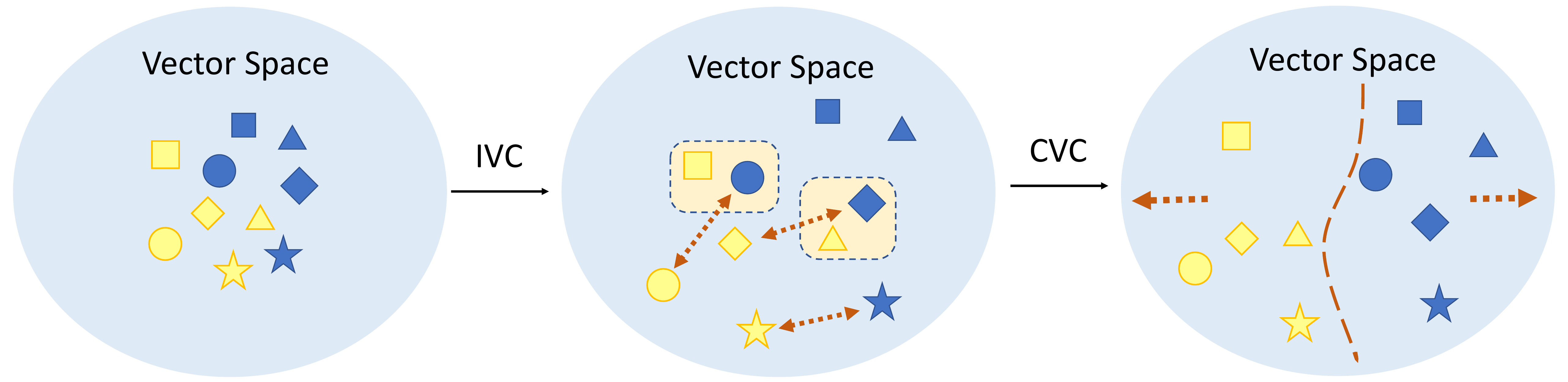}
    \caption{FGC contains two objectives, i.e., IVC and CVC. IVC pushes away dialogues with the same context but different responses (icons in the same shape), while dialogues that belong to different categories may still be similar. CVC further solves this problem by pulling all dialogues into two distinguishable clusters.}
    \label{fig:two_loss}
\end{figure*}

In this paper, we propose the \textbf{F}ine-\textbf{G}rained \textbf{C}ontrastive Learning method (\textbf{FGC}) for learning PLMs-based matching models. It consists of two complementary contrastive objectives: (1) an instance-view contrastive objective (IVC); and (2) a category-view contrastive objective (CVC). Figure \ref{fig:two_loss} demonstrates the joint effects of the two contrastive objectives on the space of matching representations. The IVC objective pushes away dialogue instances with the same context and different responses, making the model easier to distinguish between positive and negative responses. However, only pushing the examples with the same context away increases the risk of instances with different contexts getting closer in the representation space. As a remedy, the CVC objective further pulls all context-response pairs into two distinguishable clusters in the matching space according to whether the pair is positive or not. These two objectives are introduced in \ref{sec:ins_cl} and \ref{sec:cat_cl} respectively. For simplicity, we take BERT as an example in the following sections in the following sections. 



\subsection{Dialogue Data Augmentation}
\label{sec:augmentation}

Data augmentation takes an important role in contrastive learning \cite{zoph2020rethinking,ho2020contrastive}. 
Similar to standard contrastive learning (e.g., CERT), the first step of FGC is to create augmentations for every context-response pair. Given a context-response pair, we take an augmentation method on the response to generate an augmented response. The context and augmented response pair form the augmentation of the original context-response pair. In order to fine-grained control the difference between a dialogue and its corresponding augmentation and easily perform augmentation on various languages, a fully unsupervised rule-based utterance augmentation method is adopted for utterance augmentation. Inspired by \cite{wei-zou-2019-eda}, we adopt three types of augmentation operations:
\begin{itemize}
    \item \textbf{Random deletion}: Each token in the utterance is randomly and independently deleted with a probability $p_{del}$. 
    \item \textbf{Random swaping}: Each token in the utterance is randomly swapped with another token in the utterance with a probability $p_{swap}$.
    \item \textbf{Synonym replacing}: Randomly replace a non-stop-word token to one of its synonyms with a probability $p_{syn}$.
\end{itemize}

Given a response utterance $r$ and an augmentation strength $p \in [0,1]$, we randomly pick out one of these three augmentation methods and then apply the augmentation on the utterance with the probability being $p$. After augmentation, the response $r$ is converted into another augmented response $\bar{r}$. The augmentation strength $p$ is a hyper-parameter that controls how much difference is there between $r$ and $\bar{r}$. 

\subsection{Instance-View Contrastive Objective}
\label{sec:ins_cl}

\begin{figure}[htbp]
    \centering
    \includegraphics[width=0.5\textwidth,trim=0 0 0 0,clip]{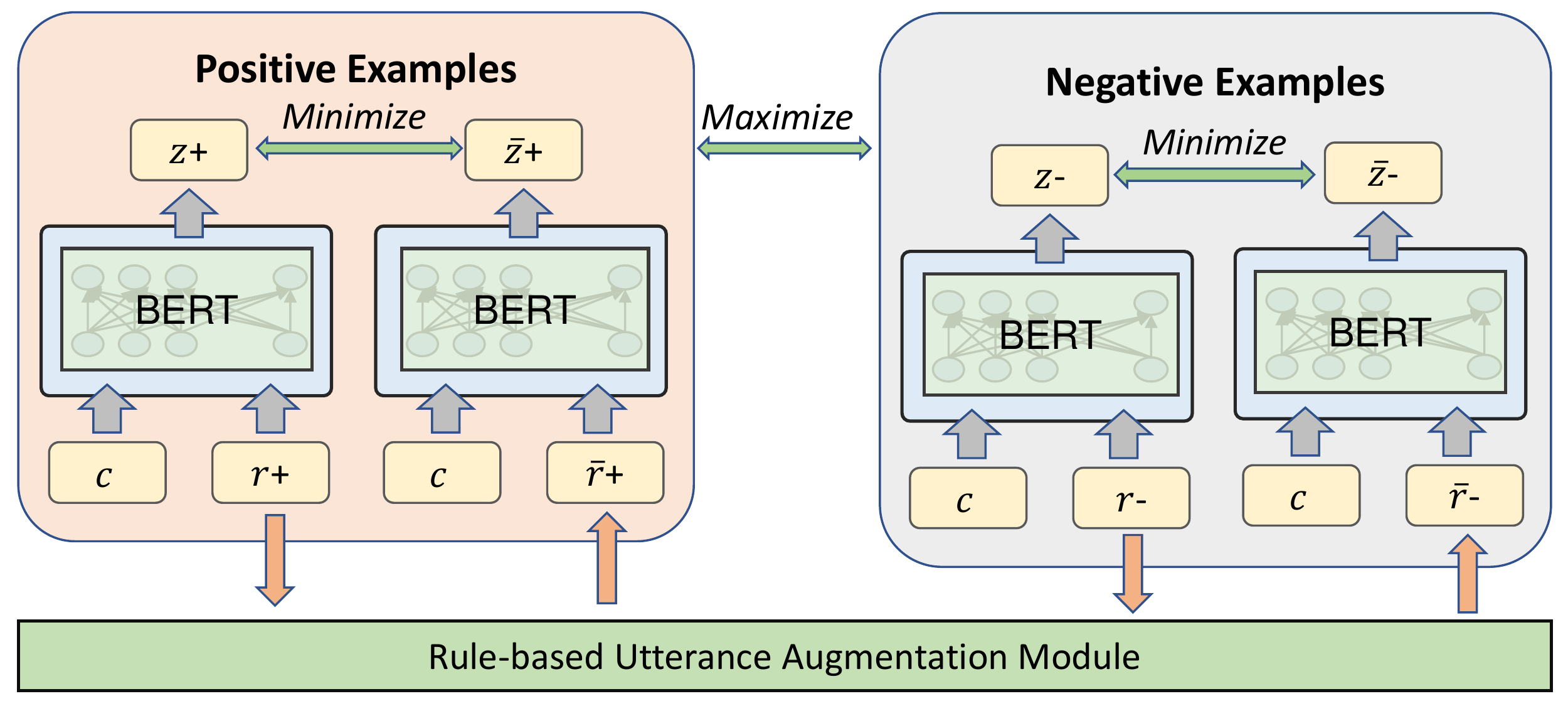}
    \caption{An overview of IVC learning. The input is a dialogue context $\mathbf{c}$ and a pair of positive and negative responses ($\mathbf{r}+$, $\mathbf{r}-$). Both responses are augmented with a rule-based utterance augmentation model to form a new pair ($\mathbf{\bar{r}}+$, $\mathbf{\bar{r}}-$). We concatenate the context $\mathbf{c}$ with four responses and fed them into the BERT encoder, which outputs a projection vector $\mathbf{z}$ for each dialogue. IVC aims to maximize the dissimilarity of $\mathbf{z}$ between positive examples and negative examples, as well as maintains high cohesion within positive and negative cases.}
    \label{fig:fine_contras}
\end{figure}


The instance-view contrastive (IVC) objective aims at introducing more discrepancy between a pair of dialogues with the same context and positive/negative responses.
Feeding a dialogue into BERT, BERT helps to make internal interactions by attention mechanism and generate latent vectors representing the dialogue. The output vector of the $\texttt{[CLS]}$ position $\mathbf{s}_{\texttt{[CLS]}}$ stands for a aggregated sequence representation of both context and response. We also take this vector as the dialogue matching representation used for contrastive learning. Moreover, we apply another projection layer to convert $\mathbf{s}_{\texttt{[CLS]}}$ into a smaller vector $\mathbf{z}$. This projection is made through an MLP with one hidden layer.
Through this projection, each coherent dialogue with positive responses $(c_i, r_i+)$ is transformed into a projection vector $\mathbf{z}_i+$, and each incoherent dialogue $(c_i, r_i-)$ is transformed into $\mathbf{z}_i-$. The augmentations of the positive and negative dialogues are also converted into two vectors, i.e.,  $\mathbf{\bar{z}}_i+$ and $\mathbf{\bar{z}}_i-$. Here $+$ and $-$ indicates the item belongs to the positive class or the negative class, and the bar indicates this item comes from an augmented example. 

As illustrated by \citet{ethayarajh-2019-contextual} and \citet{li-etal-2020-sentence}, the embedding vectors of different utterances are distributed in a narrow cone of the vector space, showing less distinguishability. This phenomenon is even worse when two utterances are semantically similar, e.g., two dialogues sharing the same context. Thus, we leverage the IVC objective on these projection vectors $\mathbf{z}$ to distinguish between positive and negative responses given the same context. IVC objective regards the projection vector $\mathbf{z}$ as a representation of response $r$ given context $c$. This loss is applied on the projection vector $\mathbf{z}$, which helps to maximize the similarity between a response with its augmentation given the same context, as well as minimize the similarity between each positive response and negative response pair. The maximum and minimum are achieved as a set of pair-wise comparisons, i.e.,
\begin{equation}
    \begin{aligned}
    \forall i \quad
    &\mathbf{sim} (\mathbf{z}_i+, \mathbf{\bar{z}_i+}) > \\
    &\quad \mathbf{sim}(\mathbf{z}_i+, \mathbf{z}_i-), \mathbf{sim}(\mathbf{z}_i+, \mathbf{\bar{z}}_i-) \\
    &\quad \mathbf{sim}(\mathbf{\bar{z}}_i+, \mathbf{z}_i-), \mathbf{sim}(\mathbf{\bar{z}}_i+, \mathbf{\bar{z}}_i-) \\
    \forall i \quad
    &\mathbf{sim} (\mathbf{z}_i-, \mathbf{\bar{z}-}_i) > \\
    &\quad \mathbf{sim}(\mathbf{z}_i-, \mathbf{z}_i+), \mathbf{sim}(\mathbf{z}_i-, \mathbf{\bar{z}}_i+) \\
    &\quad \mathbf{sim}(\mathbf{\bar{z}}_i-, \mathbf{z}_i+), \mathbf{sim}(\mathbf{\bar{z}}_i-, \mathbf{\bar{z}}_i+).
    \end{aligned}
\end{equation}
Here we use the NT-Xent Loss \cite{chen2020simple} to model the similarities of projection vectors. By writing this pair-wise comparison into a loss function, the IVC loss is formulated as 
\begin{equation}
    \begin{aligned}
    \!\! l (\mathbf{z}, \mathbf{\bar{z}}) &= -\log \frac{\exp (\mathbf{sim} (\mathbf{z}, \mathbf{\bar{z}}) / \tau)}{\sum\limits_{\mathbf{z}_k \neq \mathbf{z}} \exp (\mathbf{sim} (\mathbf{z}, \mathbf{z}_k) / \tau)} \\
    \!\! \mathcal{L}_{ivc} &= \sum_{i=1}^{N} \left( l\left(\mathbf{z}_i+, \mathbf{\bar{z}}_i+\right) + l\left(\mathbf{z}_i-, \mathbf{\bar{z}}_i-\right) \right),
    \end{aligned}
\end{equation} where $\tau > 0$ is an adjustable scalar temperature parameter that controls the separation of positive and negative classes; $\mathbf{z}_k$ ranges from $\{\mathbf{z}+, \mathbf{\bar{z}+}, \mathbf{z}-, \mathbf{\bar{z}}-\}$; and $N$ is the total number of dialogues.

Notice that the IVC objective aims to separate the representation of positive and negative responses given the same context, so that we do not take all other in-batch examples as negative examples in the same way as in standard contrastive learning.

\subsection{Category-View Contrastive Objective}
\label{sec:cat_cl}

The IVC objective ensures a high difference between dialogues with the same context, while it cannot guarantee that the learned representations are suitable for classification. The representations of a positive dialogue may be close to the representation of another negative dialogue with a different context, as is shown in Figure \ref{fig:two_loss}. Thus, we introduce another category-view contrastive (CVC) objective into model training. The category-view contrastive objective aims at bunching dialogues that belong to the same category into a cluster and separate these two clusters.

The CVC objective is applied between dialogues from the two classes. It captures the similarity of projection vectors of the same class and contrasts them with projection vectors from the other class, i.e.,
\begin{equation}
    \begin{aligned}
    & \!\! \forall{i,j,k,l} \ \ \mathbf{sim}(\mathbf{z}_i+, \mathbf{z}_j+) > \mathbf{sim}(\mathbf{z}_k+, \mathbf{z}_l-) \\
    & \!\! \forall{i,j,k,l} \ \ \mathbf{sim}(\mathbf{z}_i-, \mathbf{z}_j-) > \mathbf{sim}(\mathbf{z}_k+, \mathbf{z}_l-). \\
    \end{aligned}
\end{equation}

This category-view contrastive loss works with a batch of representation vectors of size $2N$, where the number of both positive examples and negative examples is $N$. Denote $\{\mathbf{z}_1, \mathbf{z}_2, \cdots, \mathbf{z}_{2N-1}, \mathbf{z}_{2N}\}$ to be all representation vectors in a batch, where $\{\mathbf{z}_1, \mathbf{z}_2, \cdots, \mathbf{z}_N\}$ are representation vectors for positive dialogues and their augmentations, and $\{\mathbf{z}_{N+1}, \mathbf{z}_{N+2}, \cdots, \mathbf{z}_{2N}\}$ are representation vectors for negative dialogues and their representations. The CVC objective works as an additional restriction to punish the high similarity between positive-negative pairs and low similarity within all positive and negative dialogues. The following formulas give this loss:
\begin{equation}
    \begin{aligned}
    l(\mathbf{z}_i,\mathbf{z}_j) &= \log \frac{\exp (\mathbf{z}_i \cdot \mathbf{z}_j / \tau)}{\sum_{i \neq r} \exp (\mathbf{z}_i \cdot \mathbf{z}_r / \tau)} \\
    \mathcal{L}_{cvc} &= - \frac{1}{N-1} \sum_{i=1}^{2N} \sum_{i \neq j} \mathbf{1}_{\bar{y}_i=\bar{y}_j} l(\mathbf{z}_i,\mathbf{z}_j) \\
    \end{aligned}
\end{equation}.

Finally, the BERT model is fine-tuned with the standard response selection loss $\mathcal{L}_{\texttt{sel}}$ and both IVC and CVC loss. A weighted summation is computed as
\begin{equation}
    \mathcal{L} = \mathcal{L}_{\texttt{sel}} + \lambda (\mathcal{L}_{ivc} + \mathcal{L}_{cvc}),
\end{equation} where $\lambda$ is a hyper-parameter that controls the balance between response selection loss and contrastive loss. The model is optimized by minimizing the overall loss value.

\section{Experiments}

\subsection{Dataset}

\begin{itemize}
    \item \textbf{Ubuntu Dialogue Corpus V1}\\ The Ubuntu Dialogue Corpus V1 \cite{lowe2015ubuntu} is a domain-specific multi-turn conversation dataset. Conversations in this dataset are dumped from the multi-party chat room whose topic is the Ubuntu operating system. We conducted experiments on a preprocessed data released by \citet{xu2019bert}, in which numbers, URLs, and system paths are masked by placeholders. Negative responses for each dialogue are randomly sampled from other dialogues.
    \item \textbf{Douban Corpus}\\ The Douban Corpus\cite{wu2016sequential} is a Chinese dataset collected from an online social network website named Douban. Douban Corpus is an open-domain conversation corpus, whose topic is much wider than that of Ubuntu Corpus.
\end{itemize}

The statistics of these two datasets are shown in Table \ref{table:statis}. These two datasets vary greatly in both language and topic. Following previous works, we take $R_{10}@k$s as evaluation metrics, which measures the probability of having the positive response in the top $k$ ranked responses. We take $k=\{1,2,5\}$ for model evaluation.

\begin{table}[h]
\centering
\scalebox{0.9}{
\begin{tabular}{l | ccc | ccc}
\toprule
\multirow{2}{*}{Dataset} & \multicolumn{3}{c|}{Ubuntu}  & \multicolumn{3}{c}{Douban} \\
                         & Train & Val   & Test  & Train & Val  & Test  \\
\midrule
\# dialogues             & 1M    & 500K  & 500K  & 1M    & 50K  & 6670    \\
\#pos:\#neg              & 1:1   & 1:9   & 1:9   & 1:1   & 1:1  & 1.2:8.8 \\
\# avg turns             & 10.13 & 10.11 & 10.11 & 6.69  & 6.75 & 6.45    \\
\bottomrule
\end{tabular}
}
\caption{Statistics of two datasets.}
\label{table:statis}
\end{table}

\subsection{Baseline Methods}

We introduce FGC into several open-sourced PLM-based models, including BERT and ELECTRA. We also test the effectiveness of FGC on variants of BERT model, including BERT-small (H=4, L=4, H=512), BERT with domain-adaptive post training named BERT-DPT~\cite{whang2020domain}, and BERT with self-supervised tasks named BERT-UMS~\cite{whang2021ums}. Several non-PLM-based models are also compared with our proposed FGC. \footnote{A pre-trained Chinese BERT-Small is not available, thus we do not conduct experiments on it.}

\begin{table*}[htbp]
\centering
\scalebox{0.9}{

\begin{tabular}{lccccccccc}
\toprule
\multicolumn{1}{c|}{\multirow{2}{*}{Models}}& \multicolumn{3}{c|}{Ubuntu}                & \multicolumn{6}{c}{Douban}                    \\
\multicolumn{1}{c|}{}                       & R$_{10}$@1 & R$_{10}$@2 & \multicolumn{1}{c|}{R$_{10}$@5} & MAP   & MRR   & P@1   & R$_{10}$@1 & R$_{10}$@2 & R$_{10}$@5 \\ \hline \hline
\multicolumn{10}{c}{\textbf{non-PLM-based methods}}                                                                                               \\ \hline
\multicolumn{1}{l|}{Multi-View~\cite{zhou2016multi}}             & 0.662 & 0.801 & \multicolumn{1}{c|}{0.951} & 0.505 & 0.543 & 0.342 & 0.292 & 0.350 & 0.729 \\
\multicolumn{1}{l|}{SMN~\cite{wu2016sequential}}                    & 0.726 & 0.847 & \multicolumn{1}{c|}{0.961} & 0.529 & 0.569 & 0.397 & 0.233 & 0.396 & 0.724 \\
\multicolumn{1}{l|}{DUA~\cite{zhang2018modeling}}                    & 0.752 & 0.868 & \multicolumn{1}{c|}{0.961} & 0.551 & 0.599 & 0.421 & 0.243 & 0.421 & 0.780 \\
\multicolumn{1}{l|}{DAM~\cite{zhou2018multi}}                    & 0.767 & 0.874 & \multicolumn{1}{c|}{0.961} & 0.550 & 0.601 & 0.427 & 0.254 & 0.410 & 0.757 \\
\multicolumn{1}{l|}{MRFN~\cite{tao2019multi}}                   & 0.786 & 0.886 & \multicolumn{1}{c|}{0.976} & 0.571 & 0.617 & 0.448 & 0.276 & 0.435 & 0.783 \\
\multicolumn{1}{l|}{IoI~\cite{tao2019one}}                    & 0.796 & 0.894 & \multicolumn{1}{c|}{0.974} & 0.573 & 0.621 & 0.444 & 0.269 & 0.451 & 0.786 \\
\multicolumn{1}{l|}{IMN~\cite{gu2019interactive}}                    & 0.794 & 0.889 & \multicolumn{1}{c|}{0.974} & 0.576 & 0.618 & 0.441 & 0.268 & 0.458 & 0.796 \\
\multicolumn{1}{l|}{MSN~\cite{yuan2019multi}}                    & 0.800 & 0.899 & \multicolumn{1}{c|}{0.978} & 0.587 & 0.632 & 0.470 & 0.295 & 0.452 & 0.788 \\ \hline \hline
\multicolumn{10}{c}{\textbf{PLM-based Methods}}                                                                                                   \\ \hline
\multicolumn{1}{l|}{BERT}                   & 0.820 & 0.906 & \multicolumn{1}{c|}{0.978} & 0.597 & 0.634 & 0.448 & 0.279 & 0.489 & 0.823 \\
\multicolumn{1}{l|}{BERT+FGC}               & \textbf{0.829} & 0.910 & \multicolumn{1}{c|}{0.980} & \textbf{0.614} & \textbf{0.653} & \textbf{0.495} & \textbf{0.312} & 0.495 & \textbf{0.850} \\ \hline
\multicolumn{1}{l|}{BERT-DPT~\cite{whang2020domain}}               & 0.862 & 0.935 & \multicolumn{1}{c|}{0.987} & 0.609 & 0.645 & 0.463 & 0.290 & 0.505 & 0.838 \\
\multicolumn{1}{l|}{BERT-DPT+FGC}           & \textbf{0.881} & \textbf{0.945} & \multicolumn{1}{c|}{\textbf{0.990}} & 0.620 & \textbf{0.660} & \textbf{0.495} & \textbf{0.322} & 0.495 & \textbf{0.850} \\ \hline
\multicolumn{1}{l|}{BERT-UMS~\cite{whang2021ums}}               & 0.875 & 0.942 & \multicolumn{1}{c|}{0.988} & 0.625 & 0.664 & 0.499 & 0.318 & 0.482 & 0.858 \\
\multicolumn{1}{l|}{BERT-UMS+FGC}           & \textbf{0.886} & \textbf{0.948} & \multicolumn{1}{c|}{0.990} & 0.627 & 0.670 & 0.500 & \textbf{0.326} & \textbf{0.512} & \textbf{0.869} \\ \hline
\multicolumn{1}{l|}{ELECTRA}                & 0.826 & 0.908 & \multicolumn{1}{c|}{0.978} & 0.602 & 0.642 & 0.465 & 0.287 & 0.483 & 0.839 \\
\multicolumn{1}{l|}{ELECTRA+FGC}            & \textbf{0.832} & 0.912 & \multicolumn{1}{c|}{0.980} & \textbf{0.625} & \textbf{0.668} & \textbf{0.499} & \textbf{0.313} & \textbf{0.502} & \textbf{0.850} \\ \hline
\multicolumn{1}{l|}{BERT-Small}             & 0.792 & 0.888 & \multicolumn{1}{c|}{0.972} & N/A   & N/A   & N/A   & N/A   & N/A   & N/A   \\
\multicolumn{1}{l|}{BERT-Small+FGC}         & \textbf{0.800} & 0.890 & \multicolumn{1}{c|}{0.974} & N/A   & N/A   & N/A   & N/A   & N/A   & N/A   \\
\bottomrule
\end{tabular}

}
\caption{Evaluation results on the two data sets. Numbers in bold indicate that the PLM-based models using FGC outperforms the original models with a significance level $p$-value $<0.05$.} 
\label{table:result}
\end{table*}

\subsection{Implementation Details}

All models are implemented based on Pytorch and Huggingface's implementation. Each PLM model is trained for 5 epochs with a learning rate beginning from 3e-6 to 0 with a linear learning rate decay. Our model is trained with 8 Nvidia Tesla A100 GPUs, which have 40GB of memory for each of them. For more training details, please refer to Appendix \ref{appendix:details}.

\subsection{Experimental Results}

The comparison between PLMs and FGC-enhanced PLMs is shown in Table \ref{table:result}. It can be seen from the table that all PLM-based methods outperform non-PLM-based methods. By adding our proposed FGC into PLM-based models, the performance of all models is significantly improved. The maximum improvement of a standard-sized BERT for the two datasets are 1.9\% and 3.2\% respectively in terms of R$_{10}$@1. The average performance improvement also achieves 1.1\% and 2.2\%. Besides, our proposed method can also enhance the current state-of-the-art method BERT-UMS by 1.1\% and 0.8\% on two datasets in terms of R$_{10}$@1. In addition to a standard-sized BERT model, we also find an absolute gain of 0.9\% by adding FGC on the BERT-Small model, which is about 10$\times$ smaller than a standard one. The success of these two datasets demonstrates the effectiveness of our proposed FGC across different models, languages, and dialogue topics on multi-turn response selection. 

FGC separates representation vectors of dialogues into different latent spaces according to their type of relevance between contexts and responses. On the one hand, IVC helps distinguish between positive and negative responses given the same context. On the other hand, CVC separates representations of dialogues from two categories so that these representations can have better distinguishability. As a result, the matching representation of context-response pairs for positive and negative responses are forced to stay away from each other. This better representation ensures higher accuracy in selecting the positive response given a candidate set of responses.

\section{Closer Analysis}

We conduct closer analysis with BERT-DPT since combining post-training and fine-tuning is the most popular manner of applying BERT for down-streaming tasks. The Ubuntu Corpus is used in the following analysis.

\subsection{Ablation Studies}

\begin{table}[htbp]
\centering
\scalebox{0.95}{
\begin{tabular}{lccc}
\toprule
Strategy        & R$_{10}$@1    & R$_{10}$@2    & R$_{10}$@5    \\ \hline
BERT-DPT + FGC  & \textbf{0.881} & \textbf{0.944} & \textbf{0.990} \\
\quad - IVC     & 0.866 & 0.935 & 0.986 \\
\quad - CVC     & 0.877 & 0.941 & 0.988 \\
BERT-DPT        & 0.862 & 0.935 & 0.987 \\
\bottomrule
\end{tabular}
}
\caption{Ablation Analysis on the Ubuntu corpus.}
\label{tab:ablation}
\end{table}

As we add two contrastive learning objectives into training for response selection, we test the gain of each objective. The results are shown in Table \ref{tab:ablation}. It can be observed from the table that both IVC and CVC can enhance the performance on response selection, with an absolute improvement of 1.4\% and 0.4\% respectively in terms of R$_{10}$@1. By applying these two contrastive objectives, we obtain an absolute improvement of 1.9\% based on the post-trained BERT model. Both of the two contrastive objectives share the same purpose of separating the representation of dialogues with positive and negative responses, and thus there is a performance overlap by adding these two objectives. 

\subsection{Sensitive Analysis}

\paragraph{Temperature} Temperature $\tau$ works as a hyperparameter that controls the punishment on the degree of separation of positive and negative classes. A smaller $\tau$ gives more power to pull away dialogues from different classes. We test how this hyperparameter can influence the response selection performance. We test $\tau$ in the range of $\{0.1, 0.5, 1\}$ on FGC and the results are shown in Table \ref{table:temperature}. FGC achieves the best performance when $\tau$ is set to be 0.5, while the performance drops given a smaller or a bigger $\tau$. A suitable $\tau$ can provide a proper differentiation that is neither too strong nor too weak, keeping a balance between contrastive and response selection objectives. 

\begin{table}[htbp]
\centering
\scalebox{0.95}{
\begin{tabular}{lccc}
\toprule
Temperature $\tau$  & R$_{10}$@1    & R$_{10}$@2    & R$_{10}$@5    \\ \hline
BERT-DPT      & 0.862 & 0.935 & 0.987 \\
\quad + FGC ($\tau$=0.1)          & 0.872 & 0.939 & \textbf{0.990} \\
\quad + FGC ($\tau$=0.5)          & \textbf{0.881} & \textbf{0.944} & \textbf{0.990} \\
\quad + FGC ($\tau$=1.0)          & 0.876 & 0.938 & \textbf{0.990} \\
\bottomrule
\end{tabular}
}
\caption{Influence of temperature $\tau$ in FGC.}
\label{table:temperature}
\end{table}

\paragraph{Utterance Augmentation Strength}
Utterance augmentation plays an important role in contrastive learning. A dialogue with a context and a positive response is drawn closer to its augmentation while pushed far away from the dialogue with the same context but a negative response. The strength of utterance augmentation decides the boundary of each cluster. We conduct experiments to test how augmentation strength can influence response selection accuracy. We range the augmentation strength $p$ from $\{0.1, 0.2, 0.5\}$, and the testing results are shown in Table \ref{table:augment_strength}. It achieves the best performance when $p$ equals 0.2. Augmentation strength being either too large or too small may harm the clustering. On the one hand, a too-large $p$ brings too much noise into the clustering process, which blurs the boundary between positive and negative examples. On the other hand, a too-small $p$ cannot provide enough variation to the utterance, which harms the generalization of identifying positive and negative responses.

\begin{table}[htbp]
\centering
\scalebox{0.95}{
\begin{tabular}{lccc}
\toprule
Augment Strength $p$  & R$_{10}$@1    & R$_{10}$@2    & R$_{10}$@5    \\ \hline
BERT-DPT      & 0.862 & 0.935 & 0.987 \\
\quad + FGC ($p$=0.1)          & 0.874  & 0.938  & 0.989 \\
\quad + FGC ($p$=0.2)          & \textbf{0.881} & \textbf{0.944} & \textbf{0.990} \\
\quad + FGC ($p$=0.5)          & 0.872  & 0.935  & \textbf{0.990} \\
\bottomrule
\end{tabular}
}
\caption{Influence of utterance augmentation strength $p$ in FGC.}
\label{table:augment_strength}
\end{table}

\subsection{Discussion}

\paragraph{Compare with Standard Contrastive Learning}
\label{exp:simple_contrast}

The main difference between our proposed FGC and standard contrastive learning (e.g., CERT~\cite{fang2020cert} and SimCSE~\cite{gao2021simcse}) is that we only take dialogues with the same context but different responses as negative examples, instead of using in-batch examples as negative ones. We compare FGC with those methods, whose results are shown in Table \ref{tab:compare}. Standard contrastive learning can bring less gain (or even harm) on the response selection task, while contrastive learning with fine-grained negative examples leads to a significant gain on this task.

\begin{table}[htbp]
\centering
\scalebox{0.95}{
\begin{tabular}{lccc}
\toprule
Contrastive Method  & R$_{10}$@1    & R$_{10}$@2    & R$_{10}$@5    \\ \hline
BERT-DPT            & 0.862 & 0.935 & 0.987 \\
BERT-DPT + CERT     & 0.855  & 0.931  & 0.985 \\
BERT-DPT + SimCSE   & 0.864 & 0.936 & 0.987 \\
BERT-DPT + FGC      & \textbf{0.881} & \textbf{0.944} & \textbf{0.990} \\
\bottomrule
\end{tabular}
}
\caption{Influence of utterance augmentation strength $p$ in FGC.}
\label{tab:compare}
\end{table}

\paragraph{Similarity between Dialogues}

\begin{table}[htbp]
\centering
\scalebox{0.95}{
\begin{tabular}{lcc}
\toprule
Strategy        & Ins Sim   & Cat Sim\\ \hline
BERT-DPT        & +0.074    & +0.064 \\
\quad + IVC     & -0.178    & -0.015 \\
\quad + CVC     & +0.052    & -0.109  \\
BERT-DPT + FGC  & -0.111    & -0.131 \\
\bottomrule
\end{tabular}
}
\caption{Similarity Analysis on the Ubuntu corpus.}
\label{tab:similarity}
\end{table}

The goal of FGC is to enlarge distances between dialogue examples with the same context and different responses. To estimate how effective this target is achieved, we compute two average cosine similarities: (1) instance-level similarity, which is the average similarity between dialogue pairs with the same context but different responses; and (2) category-level similarity, which is the average similarity between all positive dialogues and negative dialogues. As can be seen from Table \ref{tab:similarity}, both similarities are lowered from a positive value indicating positive correlation into a negative value indicating negative correlation by adding FGC. By introducing better distinguishability into dialogue representations, our proposed FGC helps to make better response predictions effectively. Though these two similarities can also be lowered by adding IVC alone, the category similarity is not small enough to separate the two categories well. This shortcoming is compensated by further applying CVC as an additional training objective.
Besides, CVC alone can neither provide a sufficiently low level of instance-level similarity that separates dialogues with the same context.

\paragraph{Effect of Data Augmentation Alone}
Data augmentation, working as a kind of data noise, shows a positive effect on training models with robustness in natural language processing. One may concern that can data augmentation alone help with the response selection task. We conducted experiments with data augmentation alone, i.e., no contrastive learning strategy is included. The results are shown in Table \ref{tab:augmentation_alone}. It can be observed from the table that data augmentation alone cannot enhance the model but even harm the accuracy significantly.
Data augmentation methods should work with fine-grained contrastive learning to make positive effects for the multi-turn response selection task.

\begin{table}[htbp]
\centering
\scalebox{0.95}{
\begin{tabular}{l|ll}
\toprule
             & Ubuntu & Douban \\ \hline
BERT-DPT     & 0.862  & 0.290  \\
\quad +Aug & 0.837 (-2.5\%) & 0.278 (-1.2\%)    \\
BERT-UMS     & 0.875  & 0.318  \\
\quad +Aug & 0.851 (-2.4\%) & 0.292 (-2.6\%)\\
\bottomrule
\end{tabular}
}
\caption{Model performance with data augmentation alone.}
\label{tab:augmentation_alone}
\end{table}
\section{Conclusion}

In this paper, we propose FGC, a fine-grained contrastive learning method, which helps to improve the multi-turn response selection task with PLM-based models. FGC consists of an instance-view contrastive (IVC) objective that helps to differentiate positive response and negative response with the same context, and a category-view contrastive (CVC) objective that separate positive dialogues and negative dialogues into two distinguishable clusters. Experiments and analysis on two benchmark datasets and five PLM-based models demonstrates the effectiveness of FGC to significantly improve the performance of multi-turn dialogue response selection.

\bibliography{aaai22}

\clearpage
\appendix
\section{More Implementation Details}
\label{appendix:details}

\paragraph{Domain-adaptive Post-training} For domain-adaptive post-training, we take the same hyper-parameter settings as BERT-DPT~\cite{whang2020domain}. Concretely, the maximum length of input dialogue is set to be 512. 
A full dialogue is randomly cut into a shorted token sequence with a probability of 10\%. 
A masked language model loss and a next sentence prediction loss is optimized jointly during post-training. 
For the masked language model training, we masked each token with a probability of 15\%. The post-training process traverses all the dialogues for 10 iterations, and the words that are masked during each iteration are independently sampled. 

\paragraph{Fine-tuning for Response Selection} The model is fine-tuned with the response selection task. The projection layer for transforming [CLS] vectors into projection vectors $z$ is an MLP with one hidden layer with hidden size being 256. For dialogues longer than 512 (i.e. the maximum length supported by BERT), we discard the beginning of its context while keeps a complete response, as the latter part of the dialogue context may have stronger relevance with the response. We take an AdamW optimizer\cite{loshchilov2017decoupled} with linear learning rate decay for fine-tuning. The initial learning rate is $3*10^{-5}$, and gradually decreases to 0 within 5 epochs. The $\lambda$ for controlling the balance between response selection loss and contrastive loss is set to be 1.

All pre-trained language model checkpoints are downloaded from huggingface\footnote{https://huggingface.co/}, with their names as the keys except for BERT-Small. For the BERT-Small model, the pre-trained model checkpoint is downloaded  with model name ``prajjwal1/bert-small''.
Each model is trained by 3 times, and the best results among them are reported.

\end{document}